\documentclass[sigconf, screen, nonacm=true]{acmart}

\usepackage[english]{babel}
\usepackage[utf8]{inputenc} 
\usepackage[T1]{fontenc}    
\usepackage{hyperref}       
\usepackage{url}            
\usepackage{booktabs}       
\usepackage{multirow}
\usepackage{amsmath}  
\usepackage{amsfonts}       
\usepackage{nicefrac}       
\usepackage{microtype}      
\usepackage{xcolor}         
\usepackage{cleveref}
\usepackage{csquotes}
\usepackage[htt]{hyphenat}
\usepackage{graphicx}  
\usepackage{ulem}  
\usepackage{caption}
\usepackage{subcaption}
\usepackage{adjustbox}
\usepackage{xparse}
\usepackage{soul}        
\usepackage{siunitx} 

\title{A Systematic Analysis of Chunking Strategies for Reliable Question Answering}

\author{Sofia Bennani}
\authornote{Work done while at Artefact.}
\affiliation{%
	\institution{École polytechnique}
	\city{Palaiseau}
	\country{France}
}
\email{{name.surname}@polytechnique.edu}

\author{Charles Moslonka}
\orcid{0000-0002-8719-1510}
\affiliation{%
	\institution{Artefact Research Center}
	\city{Paris}
	\country{France}
}
\affiliation{%
	\institution{MICS, CentraleSupélec, Université Paris-Saclay}
	\city{Gif-sur-Yvette}
	\country{France}
}

\begin{document}
\begin{abstract}
We study how document chunking choices impact the reliability of Retrieval-Augmented Generation (RAG) systems in industry. While practice often relies on heuristics, our end-to-end evaluation on Natural Questions systematically varies chunking method (token, sentence, semantic, code), chunk size, overlap, and context length. We use a standard industrial setup: SPLADE retrieval and a Mistral-8B generator. We derive actionable lessons for cost-efficient deployment: (i) overlap provides no measurable benefit and increases indexing cost; (ii) sentence chunking is the most cost-effective method, matching semantic chunking up to $\sim 5$k tokens; (iii) a “context cliff” reduces quality beyond $\sim 2.5$k tokens; and (iv) optimal context depends on the goal (semantic quality peaks at small contexts; exact match at larger ones).
\end{abstract}
\maketitle

\section{Introduction}
Enterprises increasingly deploy RAG systems for knowledge access and user support. In industrial settings, these systems operate under strict constraints regarding latency, storage costs, and maintainability. While agentic workflows are the ultimate goal, reliability hinges on the foundational retrieval layer --- particularly how source documents are chunked.
Despite ample work on retrieval~\cite{faysseColPaliEfficientDocument2024,santhanamColBERTv2EffectiveEfficient2022,formalEffectiveEfficientSparse2024} and LLMs, chunking is often left to rules of thumb. This paper reports an end-to-end, data-driven study conducted to standardize our production defaults.
Our contributions are:
\begin{itemize}
    \item A systematic evaluation of chunking method, size, overlap, and context length for agentic question answering (QA) on Natural Questions (NQ).
    \item Compact, deployable guidance: avoid overlap; prefer sentence chunking; select context size by task; beware the “context cliff”.
    \item A reliability view that includes abstention (“NONE”) rates alongside semantic and exact-match metrics.
\end{itemize}

\section{Experimental Setup}
\begin{figure*}
    \centering
    \includegraphics[width=0.7\linewidth]{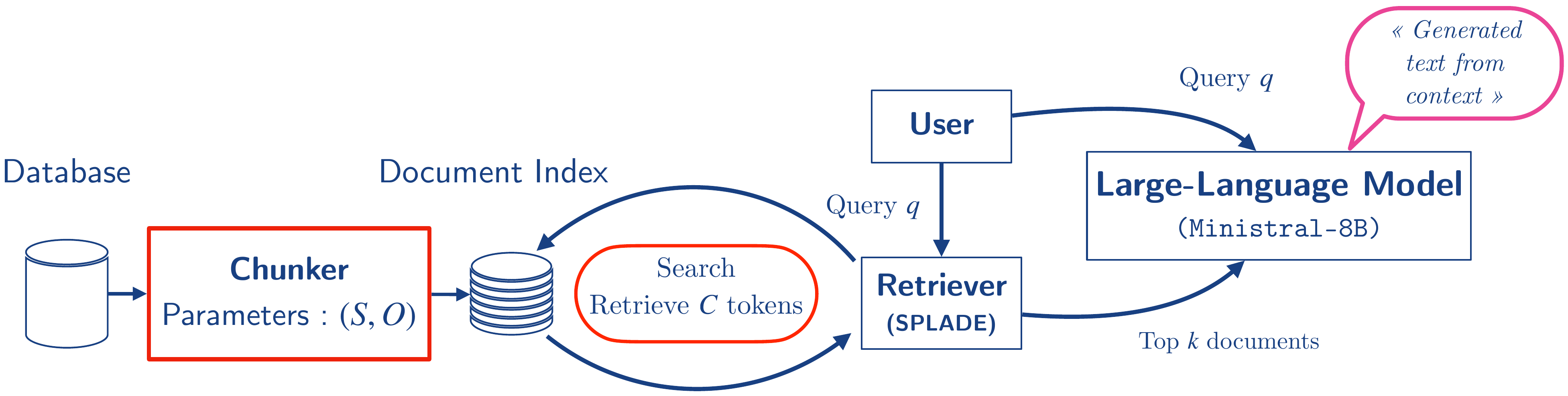}
    \caption{RAG pipeline architecture with parameters $S, O$ and $C$.}
    \label{fig:overview}
    \Description{Schematic view of the RAG pipeline, along with the different chunking parameters.}
\end{figure*}

We evaluate a standard two-stage RAG pipeline (Figure~\ref{fig:overview}): a sparse retriever indexes chunked documents; the top-ranked chunks are passed to an instruction-tuned LLM prompted to answer strictly from context and output “NONE” otherwise.

\paragraph{Task and corpus.} We use the Natural Questions~\cite{kwiatkowskiNaturalQuestionsBenchmark2019} short-answer subset (open-domain QA). The underlying corpus is English Wikipedia; documents are ingested and chunked according to each strategy below, then indexed.

\paragraph{Retrieval and context budgeting.} We use SPLADE~\cite{formalSPLADESparseLexical2021,formalSPLADEV2Sparse2021} (pretrained weights~\cite{kongSparseEmbedLearningSparse2023}) to build a sparse index over all chunks. At query time, we retrieve the top-ranked chunks and fill a token budget $C$ (context length) using the generator’s tokenizer, appending in rank order until the budget is reached. This “fill-to-budget” policy ensures fair comparison across chunk sizes and methods (no fixed-$K$ bias).

\paragraph{Chunking strategies.}
\begin{itemize}
    \item Token: fixed-size sliding windows of target size $S$ with optional token overlap $O$.
    \item Sentence: respects sentence boundaries; no sentence is split.
    \item Semantic: sentence-preserving; adjacent sentences are merged if cosine similarity (\texttt{all-MiniLM-L12-v2}) exceeds 0.5, up to the target size $S$.
    \item Code: structure-aware parsing (e.g., functions/classes) for source code, focused on markdown; included for completeness though NQ is text-centric.
\end{itemize}

\paragraph{Generation and abstention.} We use \texttt{Ministral-8B-Instruct-2410}~\cite{MinistralMinistrauxMistral} with low-temperature decoding ($T=0.1$). The prompt enforces grounded generation and explicit abstention: “Answer only using the provided context. If the context is insufficient, output ‘NONE’.” We cap output length to short answers.

\paragraph{Parameters varied.} We evaluate four \textbf{methods} (Token, Sentence, Semantic, Code) across a grid of sizes. We test \textbf{chunk sizes} $S$ from 50 to 500 (step 50), with \textbf{overlaps} $O$ of 0\% or 20\%. Finally, we retrieve into a \textbf{context budget} $C$ of \{500, 1k, 2.5k, 5k, 10k\} tokens.

\paragraph{Metrics and protocol.} We report:
\begin{itemize}
    \item BERTScore~\cite{zhangBERTScoreEvaluatingText2019} (semantic quality vs. reference answer).
    \item Exact Match (EM) with standard normalization (lowercasing; stripping punctuation/articles).
    \item None Ratio: fraction of queries where the model outputs “NONE”.
\end{itemize}
We compute 95\% bootstrap confidence intervals over questions; claims of “no measurable difference” indicate paired deltas within CIs (e.g., $|\Delta \; \mathrm{BERTScore}|\leq 0.004$, EM differences $\leq 0.001$). We intentionally do not use rerankers or LLM-as-reranker to isolate the effect of chunking and context budgeting.

\begin{figure*}
\centering
\includegraphics[width=.48\linewidth]{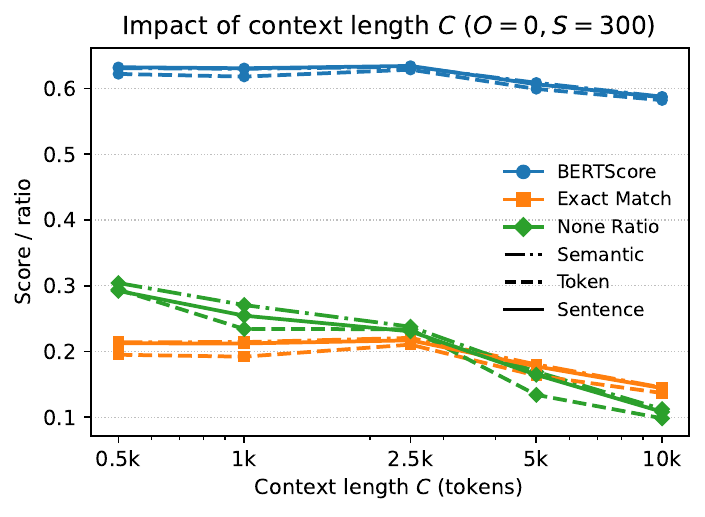}\hfill
\includegraphics[width=.48\linewidth]{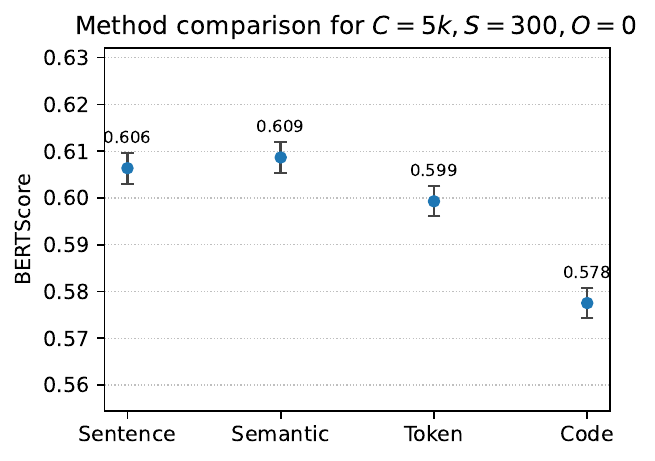}
\caption{Left: Effect of context length $C$ on metrics for different chunking methods (Sentence, Semantic, Token). Right: Chunking method comparison at fixed $C=5000$ tokens and $S=300$, $O=0$. Dots show means; bars denote 95\% bootstrap CIs.}
\Description{Two graphs. On the left one, the scores decreases with context size, and there are small but significant differences between the chunkers. The right one shows this effect more directly, showing that the sentence and semantic chunkers are quite similar. The token chunker is significantly worse that the other two, and the code chunker is so bad it's almost irrelevant.}
\label{fig:main}

\end{figure*}

\section{Findings}
We report the end-to-end effects that were most consistent and actionable across settings, along with brief mechanisms and deployment implications.

\paragraph{F1. Overlap adds cost without measurable gains.}
Across paired configurations, adding 10–20\% overlap did not improve BERTScore or EM (e.g., $|\Delta$BERTScore$|\leq 0.004$; EM differences $\leq 0.001$). Mechanism: with a sentence-aware pipeline and a sparse retriever, boundary spillover rarely changes the top-$C$ content; overlap mostly introduces near-duplicates. Cost: for overlap ratio $r$, chunk count (and index size) inflates by a factor $1/(1-r)$ (e.g., $r=0.2$ leads to $1.25\times$ more chunks), increasing ingestion time and storage. Recommendation: use $O=0$ unless you have evidence your retriever benefits from boundary redundancy.

\paragraph{F2. Method “tier list”: sentence $\approx$ semantic > token $\gg$ code (for text).}
Sentence and semantic chunking were statistically tied up to $\sim$5k tokens; token chunking lagged; code chunking was not competitive on this text task (Fig.~\ref{fig:main}). Mechanism: sentence-preserving methods keep topical coherence and reduce cross-sentence fragmentation, improving both retrieval precision and LLM grounding. Semantic merging helps when very large contexts are used (slight edge $C>5$k), likely by packing semantically contiguous text. Recommendation: default to sentence; consider semantic only for very large $C$ or highly discursive documents.

\paragraph{F3. The “context cliff”: more is not always better.}
Performance improved from small to moderate contexts but dropped beyond $\sim$2.5k tokens. For sentence chunking ($O=0$, $S=300$), BERTScore was stable between 0.5k--2.5k tokens and then declined by $\sim 4$--$5 \%$ relatively at 10k tokens. Mechanism: long-context LLMs can suffer from distraction and redundancy; retrieval at large budgets introduces overlapping or off-topic chunks, diluting signal. Recommendation: identify and enforce a sweet spot for $C$; in our setup, $C\approx 2.5$k was a strong default for QA. Note that the exact drop-off point is model-dependent; our values reflect \texttt{Ministral‑8B‑Instruct‑2410} and should be re‑tuned per LLM. However, the existence of a performance plateau or decline with excessive context is a consistent phenomenon in RAG.

\paragraph{F4. Goal-driven tuning: small $C$ for semantic quality; larger $C$ for factual accuracy; abstention is tunable.}
BERTScore tended to peak at small, focused contexts ($\sim 500$ tokens), whereas EM peaked at larger contexts ($\sim 2.5$k). None Ratio fell with larger $C$ (e.g., from $\sim 30\%$ at 0.5k to $\sim 11\%$ at 10k) and rose with larger $S$.
Mechanism: small $C$ concentrates the most relevant evidence (good for semantic faithfulness), while larger $C$ increases recall across disparate mentions (good for EM). Larger chunks reduce the number of distinct contexts retrieved, increasing abstention when narrow evidence is missed. Recommendation: for summaries/explanations, keep $C$ small; for factoid QA, use $C\approx 2.5$k. To reduce “NONE”, increase $C$ and use smaller $S$.

\begin{table*}
\centering
\caption{Practical defaults for agentic QA (text documents).}
\label{tab:defaults}
\begin{tabular}{@{}lll@{}}
\toprule
Choice & Default & Rationale \\
\midrule
Overlap $O$ & 0\% & No measurable benefit; reduces cost/complexity \\
Chunker & Sentence & Matches semantic up to $\sim$5k tokens; cheaper \\
Chunk size $S$ & 150–300 & Balances recall vs. abstention \\
Context $C$ (QA) & $\sim$2.5k & Avoids context cliff; boosts EM \\
Context $C$ (Summ.) & $\sim$500 & Maximizes semantic faithfulness \\
When $C>5$k & Consider Semantic & Slight edge at very large contexts \\
\bottomrule
\end{tabular}
\end{table*}

\paragraph{Limitations and Future Work.}
Our study focuses on optimizing the \textit{first-stage} retrieval index, a critical step for latency-sensitive industrial applications. We intentionally excluded rerankers and late-interaction models (e.g., ColBERT) to isolate the effects of chunking on the base retriever; while these methods often improve precision, they incur higher storage and latency costs that must be weighed against their benefits in future work. Furthermore, our results on Natural Questions are most representative of general text-centric corpora. While code chunking remains the clear choice for source code, these findings should be validated on specialized enterprise domains (e.g., legal or technical documentation). Finally, we used low temperature ($T=0.1$) to minimize generation variance, though key trends persisted under bootstrap resampling.

\section{Conclusion}
Chunking is a first-order design choice for reliable, cost-effective RAG agents. Our study provides compact, deployable guidance: avoid overlap; default to sentence chunking; tune context to the task; beware the context cliff beyond $\sim$2.5k tokens; and use $S$ and $C$ to control abstention. These defaults have improved the robustness of client-facing agents in practice and offer a baseline for future IR-for-agents evaluations.

\begin{acks}
    This work was done as part of the ArGiMi project, funded by BPIFrance under the France2030 national effort towards numerical common goods.
\end{acks}

\bibliographystyle{unsrt}

\end{document}